\documentclass[runningheads]{llncs}
\usepackage{graphicx}
\usepackage{comment}
\usepackage{amsmath,amssymb} 
\usepackage{color}
\usepackage{tabularx}
\usepackage{subfig}
\usepackage{algorithm}
\usepackage[noend]{algpseudocode}
\algnewcommand\algorithmicforeach{\textbf{for each}}
\algdef{S}[FOR]{ForEach}[1]{\algorithmicforeach\ #1\ \algorithmicdo}

\usepackage{bbm}

\usepackage{amsmath}
\DeclareMathOperator*{\argmin}{argmin} 
\DeclareMathOperator*{\argmax}{argmax} 

\begin{document}
\pagestyle{headings}
\mainmatter
\def\ECCVSubNumber{306}  

\title{TAFSSL: Task-Adaptive Feature Sub-Space Learning for few-shot classification} 

\titlerunning{TAFSSL}
%
\author{
Moshe Lichtenstein$^{1}$ \and
Prasanna Sattigeri$^{1}$ \and
Rogerio Feris$^{1}$ \and \\
Raja Giryes$^{2}$ \and
Leonid Karlinsky$^{1}$
}
\authorrunning{M. Lichtenstein et al.}
%
\institute{$^{1}$IBM Research AI, $^{2}$Tel Aviv University}
\maketitle

\vspace{-0.5cm}
\begin{abstract}
    The field of Few-Shot Learning (FSL), or learning from very few (typically $1$ or $5$) examples per novel class (unseen during training), has received a lot of attention and significant performance advances in the recent literature. While number of techniques have been proposed for FSL, several factors have emerged as most important for FSL performance, awarding SOTA even to the simplest of techniques. These are: the backbone architecture (bigger is better), type of pre-training on the base classes (meta-training vs regular multi-class, currently regular wins), quantity and diversity of the base classes set (the more the merrier, resulting in richer and better adaptive features), and the use of self-supervised tasks during pre-training (serving as a proxy for increasing the diversity of the base set). In this paper we propose yet another simple technique that is important for the few shot learning performance - a search for a compact feature sub-space that is discriminative for a given few-shot test task. We show that the Task-Adaptive Feature Sub-Space Learning (TAFSSL) can significantly boost the performance in FSL scenarios when some additional unlabeled data accompanies the novel few-shot task, be it either the set of unlabeled queries (transductive FSL) or some additional set of unlabeled data samples (semi-supervised FSL). Specifically, we show that on the challenging \textit{mini}ImageNet and \textit{tiered}ImageNet benchmarks, TAFSSL can improve the current state-of-the-art in both transductive and semi-supervised FSL settings by more than $5\%$, while increasing the benefit of using unlabeled data in FSL to above $10\%$ performance gain.
    \vspace{-0.1cm}
\end{abstract} 

\vspace{-0.9cm}
\section{Introduction}\label{sec:intro}
\vspace{-0.2cm}
\begin{figure}[t]
\vspace{-0.2cm}
\centering
\includegraphics[width=11cm]{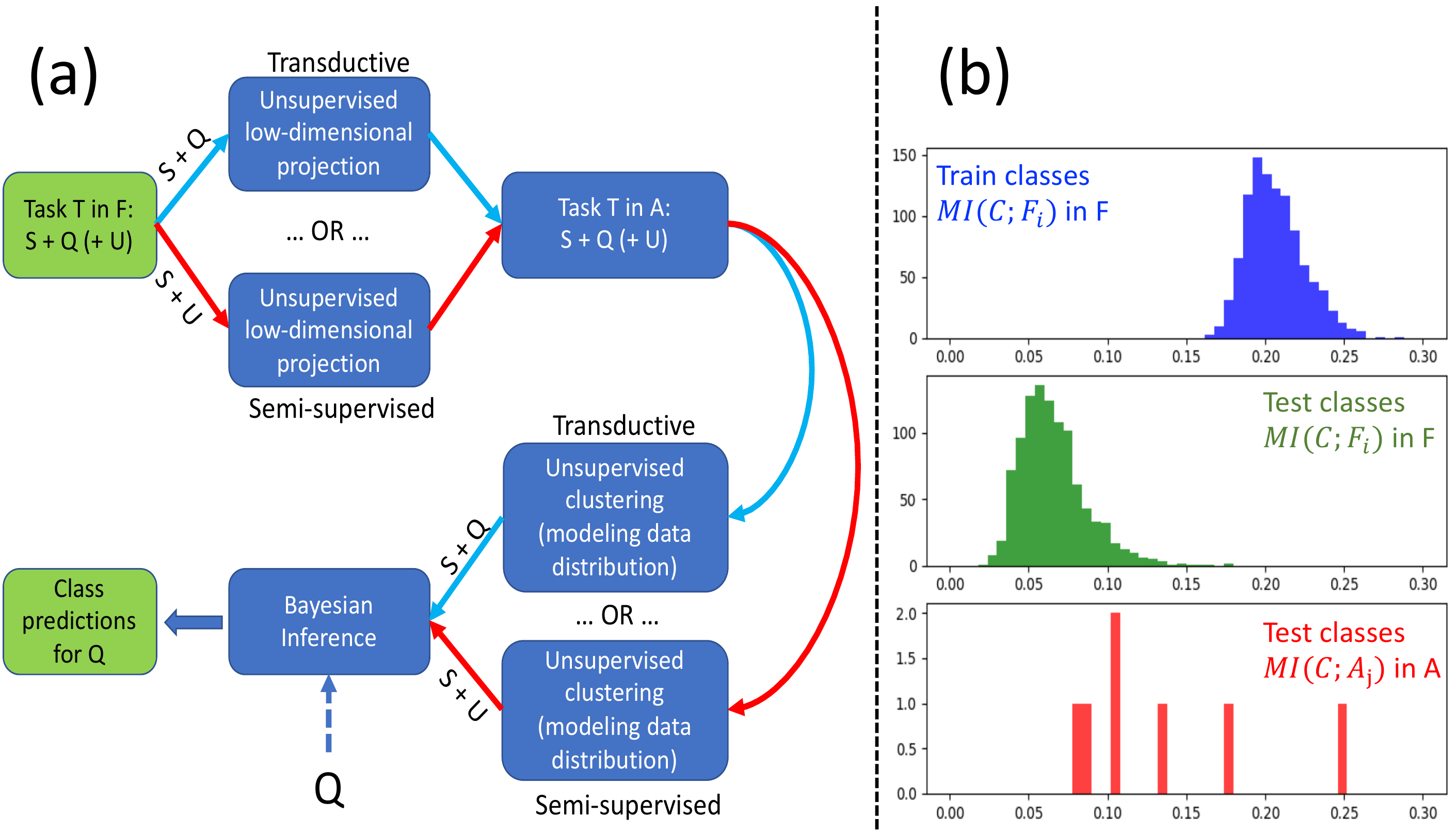}
\vspace{-0.2cm}
\caption{{\bf (a) TAFSSL overview:} red and blue pathways are for semi-supervised and transductive FSL respectively. $T$ - few-shot task; $S$ - support set; $Q$ - query set; $U$ - optional set of additional unlabeled examples (semi-supervised FSL); $F$ - original feature space; $A$ - task adapted feature sub-space. {\bf (b) Improved SNR in A:} the normalized (by min entropy) Mutual Information (MI) between either train or test classes and the features in $F$ (of dimension $1024$) or in $A$ ($7$-dim) provides the motivation to use $A$ over $F$. Computed on \textit{mini}ImageNet.}
\label{fig:overview}
\end{figure}
The great success of Deep Learning (DL) methods to solve complex computer vision problems can be attributed in part to the emergence of large labeled datasets \cite{Lin2014,imagenet} and strong parallel hardware. Yet in many practical situations, the amount of data and/or labels available for training or adapting the DL model to a new target task is prohibitively small. In extreme cases, we might be interested in learning from as little as one example per novel class. This is the typical scenario of Few-Shot Learning (FSL), a very active and exciting research topic of many concurrent works \cite{Lee2019,Snell2017,Vinyals2016,Wang2019b}. While many great techniques have been proposed to improve FSL performance, recent studies \cite{closer_look,Gidaris2019,Wang2019b} have shown that there exist a number of important factors that can improve the FSL performance, largely regardless of the model and the learning algorithm used. These include: (i) significant performance gains observed while increasing the size and the number of parameters of the backbone generating the feature representations of the images \cite{closer_look,Wang2019b}; (ii) gains while pre-training the FSL model on the base classes dataset as a regular multi-class classifier (to all base classes at once) \cite{Wang2019b}, as opposed to the popular meta-training by generating a lot of synthetic few-shot tasks from random small groups of base classes \cite{Lee2019,Vinyals2016}; (iii) gains when pre-training on more (diverse) base classes (e.g. higher empirical FSL performance on seemingly more difficult \textit{tiered}ImageNet benchmark than on supposedly simpler \textit{mini}ImageNet benchmark \cite{Lee2019,Wang2019b}; (iv) gains when artificially increasing the diversity and complexity of the base classes dataset by introducing additional self-supervised tasks during pre-training \cite{Gidaris2019}. Correctly using these factors allows the simple Nearest Neighbor classifier to attain state-of-the-art FSL performance \cite{Wang2019b} improving upon more sophisticated FSL techniques.

All the aforementioned factors and gains concern the base classes pre-training stage of the FSL methods backbones. Much less attention has been given to adapting the feature spaces resulting from these backbones to the novel classes few-shot tasks during test time. It has been shown that some moderate gains can be obtained from using the few training examples (support set) of the novel tasks to fine-tune the backbones (changing the feature spaces slightly), with best gains obtained for higher resolution and higher 'shots' (support examples per class) regimes \cite{nakamura2019}. Fine-tuning was also shown to be effective for semi-supervised setting \cite{Li2019a}, where additional unlabeled examples accompany each novel few-shot task. It has also been shown that label propagation and clustering operating in the pre-trained backbone's original feature space provide some gains for transductive FSL (allowing unlabeled queries to be used jointly to predict their labels in a bulk) and semi-supervised FSL \cite{Liu2019x,Ren2018}. Finally, meta-learned backbone architecture adaptation mechanics were proposed in \cite{metadapt} allowing for slight backbone architecture transformations adaptive to the few-shot test task.

However, slight adaptation of the backbone's feature space to a given task, using few iterations of fine-tuning on the support set or other techniques, might not be sufficient to bridge over the generalization gap introduced by the FSL backbone observing completely novel classes unseen during training (as confirmed by the relatively moderate performance gains obtained from these techniques). Intuitively, we could attribute this in part to many of the feature space dimensions (feature vector entries) becoming 'useless' for a given set of novel classes in the test few-shot task. Indeed, every feature vector entry can be seen as a certain 'pattern detector' which fires strongly when a certain visual pattern is observed on the input image. The SGD (or other) backbone training is making sure all of these patterns are discriminative for the classes used for pre-training. But, due to likely over-fitting, many of these patterns are base classes specific, and do not fire for the novel test classes. Hence, their corresponding feature vector entries will mainly produce 'noise values' corresponding to 'pattern not observed'. In other words, the ratio of feature vector entries that can be used for recognition of novel classes to ones which mainly output 'noise' significantly decreases for test few-shot task (Figure \ref{fig:overview}b). And it is unlikely that small modifications to the feature space recovers a significant portion of the 'noise producing' feature entries. The high level of noise in the feature vectors intuitively has significant adverse implications on the performance of the FSL classifier operating on this vector, especially the popular distance based classifiers like nearest-neighbor \cite{Wang2019b} and Prototypical Networks (PN) \cite{Snell2017} are affected. In light of this intuition, we conjecture that for a significant performance boost, we need to concentrate our efforts on the so-called \textit{Task-Adaptive Feature Sub-Space Learning (TAFSSL)} - seeking sub-spaces of the backbone's feature space that are discriminative for the novel classes of the test few-shot task and which are 'as noise free as possible', that is most of the sub-space dimensions indeed 'find' the patterns they represent in the images of the novel categories belonging to the task.

In this paper we set to explore TAFSSL under the transductive and the semi-supervised few-shot settings. In many practical applications of FSL, alongside the few labeled training examples (the support set) of the few shot task, additional unlabeled examples containing instances of the target novel classes are available. Such is the situation in transductive FSL which assumes that the query samples arrive in a 'bulk' and not one-by-one, and hence we can answer all the queries 'at once' while using the query set as unlabeled data. Similar situation exists in semi-supervised FSL, where unlabeled set of images simply accompanies the few-shot task. As can be observed from our experiments, TAFSSL, and especially TAFSSL accompanied by specific (proposed) forms of clustering based approaches, provides very significant boost to FSL performance under the transductive and semi-supervised setting. Specifically, we obtain over $7\%$ and over $10\%$ absolute improvement of the popular $1$-shot \textit{mini}ImageNet \cite{Vinyals2016} and \textit{tiered}ImageNet \cite{Ren2018} few-shot benchmarks in the transductive FSL setting, and $13\%$ and over $10\%$ absolute improvement in semi-supervised FSL setting (over corresponding state-of-the-art while using their respective evaluation protocols). Figure \ref{fig:overview}a illustrates an overview of the proposed approach.

To summarize, we offer the following contributions: (i) we highlight the Task-Adaptive Feature Sub-Space Learning (TAFSSL) as an important factor for Few-Shot Learning (FSL), we explore several TAFSSL methods and demonstrate significant performance gains obtained using TAFSSL for transductive and semi-supervised FSL; (ii) we propose two variants of clustering that can be used in conjunction with TAFSSL to obtain even greater performance improvements; (iii) we obtain new state-of-the-art transductive and semi-supervised FSL results on two popular benchmarks: \textit{mini}ImageNet and \textit{tiered}ImageNet; (iv) we offer an extensive ablation study of the various aspects of our approach, including sub-space dimension, unlabeled data quantity, effects of out-of-distribution noise in unlabeled data, backbone architectures, and finally - effect of class imbalance (skew) in unlabeled data (so-far unexplored in all previous works).

\vspace{-0.3cm}
\section{Related work}\label{sec:related}
\vspace{-0.2cm}
In this section we briefly review the modern Few-Shot Learning (FSL) approaches, and focus in more detail on the transductive and semi-supervised FSL methods that leverage unlabeled data.
The \textit{meta-learning} methods \cite{Vinyals2016,Snell2017,Sung2017LearningLearning,Li2019,Zhang2019} learn from few-shot tasks (or episodes) rather then from individual labeled samples. Each such task is a small dataset, with few labeled training examples (a.k.a. support), and a few test examples (a.k.a. query). The goal is to learn a model that can adapt to new tasks with novel categories unseen during training.
The \textit{gradient-based meta learners} \cite{Finn2017,Li2017,Zhou2018,Ravi2017,Munkhdalai2017,leo,closer_look,Lee2019,Oreshkin2018} search for models that are good initialization for transfer to novel few-shot tasks. Typically, in these methods higher order derivatives are used for meta-training, optimizing the loss the model would have after applying one or several gradient steps. At test time, the model is fine-tuned to the novel few shot-tasks. In \cite{Dvornik2019} ensemble methods for few-shot learning are proposed.
The \textit{few-shot learning by metric learning} methods \cite{Weinberger2009,Snell2017,Rippel2015,Garcia2017,Santoro2016,relationnet,Oreshkin2018,caml,am3,devos2019} learn a non-linear embedding into a metric space where $L_2$ nearest neighbor (or similar) is used to classify instances of new categories according to their proximity to the few support examples embedded in the same space. In \cite{devos2019,projective2019} distance to class prototype is replaced by distance to a class sub-space. As opposed to \cite{devos2019} and \cite{projective2019} that try to optimize a sub-space for each class (according to that class support examples), in TAFSSL we seek a single sub-space optimally adapted to the entire data of the few-shot task - labeled and unlabeled. Notably, in \cite{Wang2019b} regular (non-meta-learning) pre-training was used in combination with 'large' backbones (e.g. DenseNet \cite{Huang2017}) and a nearest-neighbor classifier to achieve state-of-the-art results, highlighting the importance of diverse pre-training and backbone size to FSL performance.
The \textit{generative and augmentation-based few-shot approaches} \cite{Park2015,Dosovitskiy2017,Su2015,Guu2017,cubuk2018,lim2019a,Reed2018,Antoniou2018,Hariharan2017,Schwartz2018,Wang2018,Chen2018,Yu2017,Schwartz2019,laso2019} are methods that (learn to) generate more samples from the one or a few examples available for training in a given few-shot learning task.

\textbf{Transductive and semi-supervised FSL:} In many practical applications, in addition to the labeled support set, we have additional unlabeled data accompanying the few-shot task. In transductive FSL \cite{dhillon2019,Liu2019x,Kim2019,Qiao2019} we assume the set of task queries arrives in a bulk and we can simply use it as a source of unlabeled data, allowing query samples to 'learn' from each other. In \cite{dhillon2019} the query samples are used in fine-tuning in conjunction with entropy minimization loss in order to maximize the certainty of their predictions.
In semi-supervised FSL \cite{Li2019a,Ren2018,projective2019,Liu2019x} the unlabeled data comes in addition to the support set and is assumed to have a similar distribution to the target classes (although some unrelated samples noise is also allowed). In the LST \cite{Li2019a} self-labeling and soft attention are used on the unlabeled samples intermittently with fine-tuning on the labeled and self-labeled data. Similarly to LST, \cite{Ren2018} updates the class prototypes using k-means like iterations initialized from the PN prototypes. Their method also includes down-weighting the potential distractor samples (likely not to belong to the target classes) in the unlabeled data. In \cite{projective2019} unlabeled examples are used through soft-label propagation. In \cite{Saito2019a} semi-supervised few-shot domain adaptation is considered. In \cite{Garcia2017,Liu2019x,Kim2019} graph neural networks are used for sharing information between labeled and unlabeled examples in semi-supervised \cite{Garcia2017,Liu2019x} and transductive \cite{Kim2019} FSL setting. Notably, in \cite{Liu2019x} a Graph Construction network is used to predict the task specific graph for propagating labels between samples of semi-supervised FSL task.
\vspace{-0.3cm}
\section{Method}\label{sec:method}
\vspace{-0.2cm}
In this section we derive the formal definition of TAFSSL and examine several approaches for it. In addition, we propose several ways to combine TAFSSL with clustering followed by Bayesian inference, which is shown to be very beneficial to the performance in the Results section \ref{sec:experiments}.

\vspace{-0.4cm}
\subsection{FSSL and TAFSSL}\label{sec:tafssl}
\vspace{-0.2cm}
Let a CNN backbone $\mathcal{B}$ (e.g. ResNet \cite{He2015} or DenseNet \cite{Huang2017}) pre-trained for FSL on a (large) dataset $\mathcal{D}_b$ with a set of base (training) classes $\mathcal{C}_b$. Here for simplicity, we equally refer to all different forms of pre-training proposed for FSL in the literature, be it meta-training \cite{Vinyals2016} or 'regular' training of a multi-class classifier for all the classes $\mathcal{C}_b$ \cite{Wang2019b}. Denote by $\mathcal{B}(x) \in \mathcal{F} \subset \mathbf{R}^m$ to be a feature vector corresponding to an input image $x$ represented in the feature space $\mathcal{F}$ by the backbone $\mathcal{B}$. Under this notation, we define the goal of linear Feature Sub-Space Learning (FSSL) to find an 'optimal' (for a certain task) linear sub-space $\mathcal{A}$ of $\mathcal{F}$ and a linear mapping $W$ of size $r \times m$ (typically with $r \ll m$) such that:
\begin{equation}
    \mathbf{R}^r \supset \mathcal{A} \ni A = W \cdot \mathcal{B}(x)
\end{equation}
is the new representation of an input image $x$ as a vector $A$ in the feature sub-space $\mathcal{A}$ (spanned by rows of $W$).

Now, consider an $n$-way + $k$-shot few-shot test task $\mathcal{T}$ with a query set $Q$, and a support set: 
$
    S = \{ s_i^j | 1 \le i \le n, 1 \le j \le k, \mathcal{L}(s_i^j) = i \} 
$,
where $\mathcal{L}(x)$ is the class label of image $x$, so in $S$ we have $k$ training samples (shots)
for each of the $n$ classes in the task $\mathcal{T}$. Using the PN \cite{Snell2017} paradigm we assume $k=1$ (otherwise support examples of the same class are averaged to a single class prototype) and that each $q \in Q$ is classified using Nearest Neighbor (NN) in $\mathcal{F}$: 
\begin{equation}
    CLS(q) = \argmin_{i} {|| \mathcal{B}(s_i^1) - \mathcal{B}(q) ||^2}
\end{equation}
Then, in the context of this given task $\mathcal{T}$, we can define linear Task-Adaptive FSSL (TAFSSL) as a search for a linear sub-space $\mathcal{A}_{\mathcal{T}}$ of the feature space $\mathcal{F}$ defined by a $\mathcal{T}$-specific projection matrix $W_{\mathcal{T}}$, such that the probability:
\begin{equation}\label{eq:fssl_obj}
    \frac{exp(-\tau \cdot || W_{\mathcal{T}} \cdot ( \mathcal{B}(s_{\mathcal{L}(q)}^1) - \mathcal{B}(q) ) ||^2)}{\sum_{i}{exp(-\tau \cdot || W_{\mathcal{T}} \cdot ( \mathcal{B}(s_{i}^1) - \mathcal{B}(q) ) ||^2)}}
\end{equation}
of predicting $q$ to belong to the same class as the 'correct' support $s_{\mathcal{L}(q)}^1$ is maximized, while of course the true label $\mathcal{L}(q)$ is unknown at test time (here $\tau$ in eq. \ref{eq:fssl_obj} is a temperature parameter, we used $\tau=1$). 

\vspace{-0.5cm}
\subsubsection{Discussion.} 
Using the 'pattern detectors' intuition from section \ref{sec:intro}, lets consider the activations of each dimension $F_d$ of $F \in \mathcal{F}$ as a random variable with a Mixture of (two) Gaussians (MoG) distribution:
\begin{equation}
F_d \sim P_d = \rho_n \cdot N(\mu_n, \sigma_n) + \rho_s \cdot N(\mu_s, \sigma_s)
\end{equation}
where $(\mu_n, \sigma_n)$ and $(\mu_s, \sigma_s)$ are the expectation and variance of the $F_d$'s distribution of activations when $F_d$ does not detect (noise) or detects (signal) the pattern respectively. The $\rho_n$ and $\rho_s$ are the noise and the signal prior probabilities respectively ($\rho_n + \rho_s = 1$). For brevity, we drop the index $d$ from the distribution parameters. Naturally, for the training classes $C_b$, for most dimensions $F_d$ the $\rho_s \gg 0$ implying that the dimension is 'useful' and does not produce only noise (Figure \ref{fig:overview}b, top). However, for the new (unseen during training) classes of a test task $\mathcal{T}$ this is no longer the case, and it is likely that for the majority of dimensions $\rho_s^{\mathcal{T}} \approx 0$ (Figure \ref{fig:overview}b, middle). Assuming (for the time being) that $F_d$ are conditionally independent, 
the square Euclidean distance could be seen as an aggregation of votes for the 'still useful' (for the classes of $\mathcal{T}$) patterns, and a sum of squares of i.i.d (zero mean) Gaussian samples for the patterns that are 'noise only' on the classes of $\mathcal{T}$. The latter 'noise dimensions' randomly increase the distance on the expected order of $N^{\mathcal{T},\mathcal{F}} \cdot \sigma_n^2$, where $N^{\mathcal{T},\mathcal{F}}$ is the number of \textit{noise features} of the feature space $\mathcal{F}$ for the classes of task $\mathcal{T}$. 
Using this intuition, if we could find such a TAFSSL sub-space $\mathcal{A}_{\mathcal{T}}$ adapted to the task $\mathcal{T}$ so that $N^{\mathcal{T},\mathcal{S}_{\mathcal{T}}}$ is reduced (Figure \ref{fig:overview}b, bottom), we would improve the performance of the NN classifier on $\mathcal{T}$.
%
With only few labeled samples in the support set $S$, we cannot expect to effectively learn the $W_{\mathcal{T}}$ projection to the sub-space $\mathcal{A}_{\mathcal{T}}$ using SGD on $S$. Yet, when unlabeled data accompanies the task $\mathcal{T}$ ($Q$ in transductive FSL, or an additional set of unlabeled samples $U$ in semi-supervised FSL), we can use this data to find such $W_{\mathcal{T}}$ that: (a) the dimensions of $\mathcal{A}_{\mathcal{T}}$ are 'disentangled', meaning their pairwise independence is maximized; (b) after the 'disentanglement' we choose the dimensions that are expected to 'exhibit the least noise' or in our previous MoG notation have the largest $\rho_s$ values. 

Luckily, simple classical methods can be used for TAFSSL approximating the requirements (a) and (b). Both Principle Component Analysis (PCA) \cite{pca} and Independent Component Analysis (ICA) \cite{ica} applied in $\mathcal{F}$ on the set of samples: $S \cup Q$ (transductive FSL) or $S \cup U$ (semi-supervised FSL) can approximate (a). PCA under the approximate joint Gaussianity assumption of $\mathcal{F}$, and ICA under approximate non-Gaussianity assumption. In addition, if after the PCA rotation we subtract the mean, the variance of the (zero-mean) MoG mixtures for the transformed (independent) dimensions would be:
\begin{equation}
    \rho_n \cdot (\mu_n^2 + \sigma_n^2) + \rho_s \cdot (\mu_s^2 + \sigma_s^2)
\end{equation}
Then assuming $\mu_n$ and $\sigma_n$ are roughly the same for all dimensions (which is reasonable due to heavy use of Batch Normalization (BN) in the modern backbones), choosing the dimensions with higher variance in PCA would lead to larger $\rho_s$, $\mu_s$, and $\sigma_s$ - all of which are likely to increase the signal-to-noise ratio of the NN classifier. Larger $\mu_s$ leads to patterns with stronger 'votes', larger $\sigma_s$ means wider range of values that may better discriminate multiple classes, and larger $\rho_s$ means patterns that are more frequent for classes of $\mathcal{T}$. Similarly, the dimensions with bigger $\rho_s$ exhibit stronger departure from Gaussianity and hence would be chosen by ICA. 

\vspace{-0.5cm}
\subsubsection{TAFSSL summary.}
To summarize, following the discussion above, both PCA and ICA are good simple approximations for TAFSSL using unlabeled data and therefore we simply use them to perform the 'unsupervised low-dimensional projection' in the first step of our proposed approach (Figure \ref{fig:overview}a). As we show in the Results section \ref{sec:experiments}, even on their own (when directly followed by an NN classifier) they lead to significant FSL performance boosts (Tables \ref{tab:trans} and \ref{tab:semi}).

\vspace{-0.4cm}
\subsection{Clustering}\label{sec:clustering}
\vspace{-0.2cm}
It was shown that clustering is a useful tool for transductive and semi-supervised FSL \cite{Ren2018}. There, it was assumed that modes of the task $\mathcal{T}$ data distribution (including both labeled and unlabeled image samples) correspond classes. However, in the presence of feature 'noise' in $\mathcal{F}$, as discussed is section \ref{sec:tafssl}, the 'class' modes may become mixed with the noise distribution modes, that may blur the class modes boundaries or swallow the class modes altogether. Indeed, the performance gains in \cite{Ren2018} were not very high.

In contrast, after applying PCA or ICA based TAFSSL, the feature noise levels are usually significantly reduced (Figure \ref{fig:overview}b) making the task-adapted feature sub-space $\mathcal{A}_{\mathcal{T}}$ of the original feature space $\mathcal{F}$ to be much more effective for clustering. We propose two clustering-based algorithms, the Bayesian K-Means (BKM) and Mean-Shift Propagation (MSP). In the Results section \ref{sec:experiments} we show that following PCA or ICA based TAFSSL, these clustering techniques add about $5\%$ to the performance. They are used to perform the 'unsupervised clustering' + 'bayesian inference' steps of our approach (Figure \ref{fig:overview}a).

The \textbf{BKM} is a soft k-means \cite{kmeans} variant accompanied with Bayesian inference for computing class probabilities for the queries. In BKM, each k-means cluster, obtained for the entire set of (labeled + unlabeled) task $\mathcal{T}$ data, is treated as a Gaussian mixture distribution with a mode for each class. The BKM directly computes the class probability for each query $q \in Q$ by averaging the posterior of $q$ in each of the mixtures with weights corresponding to $q$'s association probability to each cluster. The details of BKM are summarized in Algorithm \ref{algo:bkm} box.

\vspace{-0.2cm}
\begin{algorithm}[H] 
\captionof{algorithm}{Bayesian K-Means (BKM)}
\label{algo:bkm}
	\begin{algorithmic}[]
        \State Cluster the samples of task $\mathcal{T}$ ($Q \cup S$ or $U \cup S$ in transductive or semi-supervised FSL respectively) into $k$ clusters, associating each to $c_k$ - the centroid of cluster $k$. 
		\ForEach{$s \in S$, $q \in Q$, and $k$}
		\State $P(cluster(q) = k) = \frac{\exp(-||q - c_k||^2)}{\sum_j \exp(-||q - c_j||^2)}$
		\State $P(cluster(s) = k) = \frac{\exp(-||s - c_k||^2)}{\sum_j \exp(-||s-c_j||^2)}$
		\State $P(\mathcal{L}(q) = i| cluster(q) = k) = \sum_{\mathcal{L}(s)=i} \frac{\exp(-||q - s||^2)\cdot P(cluster(s) = k)}{\sum_{t \in S} \exp(-||q - t||^2) \cdot P(cluster(t) = k)}$
		\EndFor
        \State $P(\mathcal{L}(q) = i) = \sum_k P(\mathcal{L}(q) = i|cluster(q) = k) \cdot P(cluster(q) = k)$
	\end{algorithmic}
\end{algorithm} 

The \textbf{MSP} is a mean-shift \cite{meanShift} based approach, that is used to update the prototype of each class. In MSP we perform a number of mean-shift like iterations on the prototypes \cite{Snell2017} of the classes taken within the distribution of all the (labeled and unlabeled) samples of $\mathcal{T}$. In each iteration, for each prototype $p_i$ (of class $i$), we compute a set of $K$ most confident samples within a certain confidence radius and use the mean of this set as the next prototype (of class $i$). The $K$ itself is balanced among the classes. The details of MSP are summarized in Algorithm \ref{algo:msp} box. Following MSP, the updated prototypes are used in standard NN classifier fashion to obtain the class probabilities.

\vspace{-0.2cm}
\begin{algorithm}[H] 
\captionof{algorithm}{Mean-Shift Propagation (MSP)}
\label{algo:msp}
	\begin{algorithmic}[]   
		\State \textbf{Initialize:}
		\State Compute prototypes: $\{p_i = \frac{1}{k} \cdot \sum_{s \in S, \mathcal{L}(s)=i} s \}$, where $k$ is \# of shots in task $\mathcal{T}$
		\For{N times}
        \State Compute $P(\mathcal{L}(x)=i) = \frac{\exp(-||x - p_i||^2)}{\sum_j \exp(-||x - p_j||^2)}$, $\forall x \in Q \cup S$ (or $x \in U \cup S$)
        \State Compute predictions  $c(x) = \argmax_{i}{P(\mathcal{L}(x)=i)}$
		\State $K_i = \sum_{x} \mathbbm{1}_{(c(x)=i) \wedge (P(\mathcal{L}(x)=i)>T)}$, where $T$ is a threshold parameter
		\State K = $min_i\{K_i\}$
		\State Compute the new prototypes: $\{p_i = \frac{1}{K} \cdot \sum_{x \in \hat{S}_i} x \}$, where $\hat{S}_i$ are the top $K$ samples that have  $c(x) = i$ sorted in decreasing order of $P(\mathcal{L}(x)=i)$
		\EndFor
        \State \Return labels $c(q), \forall q \in Q$  
	\end{algorithmic}
\end{algorithm} 

\vspace{-0.4cm}
\subsection{Implementation details}\label{sec:impl}
\vspace{-0.2cm}
All the proposed TAFSSL approaches were implemented in PyTorch. Our code will be released upon acceptance. We have used the PyTorch native version of SVD for PCA, and FastICA from sklearn for ICA. The k-means from sklearn was used for BKM. The sub-space dimensions were $4$ for PCA based TAFSSL, and $10$ for ICA based TAFSSL. These were set using validation (section \ref{sec:dim}). The $T=0.3$ and $N=4$ were used for MSP, and $k=5$ for BKM, all set using validation. We used the backbones implementations from \cite{Wang2019b}. Unless otherwise specified, DenseNet backbone was used (for backbones ablation, please see section \ref{sec:backbones}). Using the most time consuming of the proposed TAFSSL approaches (ICA + BKM) our running time was measured as below $0.05$ seconds (CPU) for a typical $1$-shot and $5$-way episode with $15$ queries per class.
\vspace{-0.3cm}
\section{Results}\label{sec:experiments}
\vspace{-0.2cm}
We have evaluated our approach on the popular few-shot classification benchmarks, namely \textit{mini}ImageNet \cite{Vinyals2016} and \textit{tiered}ImageNet \cite{Ren2018}, used in all transductive and semi-supervised FSL works \cite{dhillon2019,Liu2019x,Kim2019,Qiao2019,Li2019a,Ren2018,projective2019,Liu2019x}. On these benchmarks, we used the standard evaluation protocols, exactly as in corresponding (compared) works. The results of the transductive and semi-supervised FSL evaluation, together with comparison to previous methods, are summarized in tables \ref{tab:trans} and \ref{tab:semi} respectively and are detailed and discussed in the following sections. All the performance numbers are given in accuracy $\%$ and the $0.95$ confidence intervals are reported. The tests are performed on $10,000$ random $5$-way episodes, with $1$ or $5$ shots (number of support examples per class), and with $15$ queries per episode (unless otherwise specified). For each dataset, the standard train / validation / test splits were used. For each dataset, training subset was used to pre-train the backbone (from scratch) as a regular multi-class classifier to all the train classes, same as in \cite{Wang2019b}; the validation data was used to select the best model along the training epochs and to choose the hyper-parameters; and episodes generated from the test data (with test categories unseen during training and validation) were used for meta-testing to obtain the final performance. In all experiments not involving BKM, the class probabilities were computed using the NN classifier to the class prototypes.

\vspace{-0.3cm}
\subsection{FSL benchmarks used in our experiments}\label{sec:datasets}
\vspace{-0.2cm}
%
The \textbf{\textit{mini}ImageNet benchmark (Mini)} \cite{Vinyals2016} is a standard benchmark for few-shot image classification, that has $100$ randomly chosen classes from ILSVRC-2012 \cite{imagenet}. They are randomly split into disjoint subsets of $64$ meta-training, $16$ meta-validation, and $20$ meta-testing classes. Each class has 600 images of size $84 \times 84$. We use the same splits as \cite{Lee2019} and prior works.

The \textbf{\textit{tiered}ImageNet benchmark (Tiered)} \cite{Ren2018} is a larger subset of ILSVRC-2012 \cite{imagenet}, consisted of $608$ classes grouped into $34$ high-level classes. These are divided into disjoint $20$ meta-training high-level classes, $6$ meta-validation classes, and $8$ meta-testing classes. This corresponds to $351$, $97$, and $160$ classes for meta-training, meta-validation, and meta-testing respectively. Splitting using higher level classes effectively minimizes the semantic similarity between classes belonging to the different splits. All images are of size $84 \times 84$.

\vspace{-0.3cm}
\subsection{Transductive FSL setting}\label{sec:exp_trans}
\vspace{-0.2cm}
In these experiments we consider the transductive FSL setting, where the set of queries is used as the source of the unlabeled data. This setting is typical for cases when an FSL classifier is submitted a bulk of query data for offline evaluation. In Table \ref{tab:trans} we report the performance of our proposed TAFSSL (PCA, ICA), clustering (BKM, MSP), and TAFSSL+clustering (PCA/ICA + BKM/MSP) approaches and compare them to a set of baselines and state-of-the-art (SOTA) transductive FSL methods from the literature: TPN \cite{Liu2019x} and Transductive Fine-Tuning \cite{dhillon2019}. We also compare to SOTA regular FSL result of \cite{Wang2019b} in order to highlight the effect of using the unlabeled queries for prediction. As baselines, we try to maximally adapt the method of \cite{Wang2019b} to the transductive FSL setting. These are the so-called "trans-mean-sub" that on each test episode subtracts the mean of all the samples ($S \cup Q$) from all the samples followed by L2 normalization (in order reduce the episode bias); and the "trans-mean-sub(*)" where we do the same but computing and subtracting the means of the $S$ and $Q$ sample sets separately (in order to better align their distributions). As can be seen from Table \ref{tab:trans}, on both the Mini and the Tiered transductive FSL benchmarks, the top performing of our proposed TAFSSL based approaches (ICA+MSP) consistently outperforms all the previous (transductive and non-transductive) SOTA and the baselines by more then $10\%$ in the more challenging $1$-shot setting and by more then $2\%$ in the $5$-shot setting, underlining the benefits of using the transductive setting, and the importance of TAFSSL to this setting. In the following section, we only evaluate the ICA based TAFSSL variants as it was found to consistently outperform the PCA based variant under all settings.
\begin{table}[!t]
\vspace{-0.5cm}
\centering
\caption{Transductive setting}
\label{tab:trans}
\scriptsize{
\begin{tabular}[h]{lcccc}
\hline
& Mini 1-shot & Mini 5-shot & Tiered 1-shot & Tiered 5-shot\\
\hline

\textbf{Simple shot \cite{Wang2019b}} & 
64.30 $\pm$ 0.20 & 81.48 $\pm$ 0.14 & 71.26 $\pm$ 0.21 & 86.59 $\pm$ 0.15 \\

\textbf{TPN \cite{Liu2019x}} & 
55.51 $\pm$ 0.86 & 69.86 $\pm$ 0.65 & 59.91 $\pm$ 0.94 & 73.30 $\pm$ 0.75 \\

\textbf{TEAM \cite{Qiao2019}} & 
60.07 $\pm$ N.A. & 75.90 $\pm$ N.A. & - & - \\

\textbf{EGNN + trans. \cite{Kim2019}} & 
- & 76.37 $\pm$ N.A. & - & 80.15 $\pm$ N.A. \\

\textbf{Trans. Fine-Tuning \cite{dhillon2019}} & 
65.73 $\pm$ 0.68 & 78.40 $\pm$ 0.52 & 73.34 $\pm$ 0.71 & 85.50 $\pm$ 0.50 \\

\textbf{Trans-mean-sub} & 
65.58 $\pm$ 0.20 & 81.45 $\pm$ 0.14 & 73.49 $\pm$ 0.21 & 86.56 $\pm$ 0.15 \\

\textbf{Trans-mean-sub(*)} & 
65.88 $\pm$ 0.20 & 82.20 $\pm$ 0.14 & 73.75 $\pm$ 0.21 & 87.16 $\pm$ 0.15 \\

\hline

\textbf{PCA} & 
70.53 $\pm$ 0.25 & 80.71 $\pm$ 0.16 & 80.07 $\pm$ 0.25 & 86.42 $\pm$ 0.17 \\

\textbf{ICA} & 
72.10 $\pm$ 0.25 & 81.85 $\pm$ 0.16 & 80.82 $\pm$ 0.25 & 86.97 $\pm$ 0.17 \\

\textbf{BKM} & 
72.05 $\pm$ 0.24 & 80.34 $\pm$ 0.17 & 79.82 $\pm$ 0.25 & 85.67 $\pm$ 0.18\\

\textbf{PCA + BKM} & 
75.11 $\pm$ 0.26 & 82.24 $\pm$ 0.17 & 83.19 $\pm$ 0.25 & 87.83 $\pm$ 0.17\\

\textbf{ICA + BKM} & 
75.79 $\pm$ 0.26 & 82.83 $\pm$ 0.16 & 83.39 $\pm$ 0.25 & 88.00 $\pm$ 0.17\\

\textbf{MSP} & 
71.39 $\pm$ 0.27 & 82.67 $\pm$ 0.15 & 76.01 $\pm$ 0.27 & 87.13 $\pm$ 0.15\\

\textbf{PCA + MSP} & 
76.31 $\pm$ 0.26 & 84.54 $\pm$ 0.14 & 84.06 $\pm$ 0.25 & 89.13 $\pm$ 0.15\\

\textbf{ICA + MSP} &
\textbf{77.06 $\pm$ 0.26} & \textbf{84.99 $\pm$ 0.14} & \textbf{84.29 $\pm$ 0.25} & \textbf{89.31 $\pm$ 0.15}\\

\hline
\end{tabular}
}
\end{table}

\vspace{-0.3cm}
\subsection{Semi-supervised FSL setting}\label{sec:exp_semi}
\vspace{-0.2cm}
In this section we evaluate our proposed approaches in the semi-supervised FSL setting. In this setting, we have an additional set of unlabeled samples $U$ that accompanies the test task $\mathcal{T}$. In $U$ we usually expect to have additional samples from the $\mathcal{T}$'s target classes distribution, possibly mixed with additional unrelated samples from some number of distracting classes (please see section \ref{sec:noise} for an ablation on this). In Table \ref{tab:semi} we summarize the performance of our proposed TAFSSL based approaches, and compare them to the SOTA semi-supervised FSL methods of \cite{Ren2018,Liu2019x,Li2019a,projective2019}. In addition, we also present results for varying number of additional unlabeled samples in $U$ (where available). As can be seen from Table \ref{tab:semi}, in the semi-supervised setting, the TAFSSL-based approaches outperform all competing methods by a large margins of over $8\%$ and $4\%$ accuracy gain in both the Mini and the Tiered benchmarks in $1$-shot and $5$-shot settings respectively. Interestingly, same as for the transductive FSL, for the semi-supervised FSL the ICA+MSP approach is the best performing.
\begin{table}[!t]
\vspace{-0.5cm}
\caption{Semi supervised setting. For clarity, results are sorted according to increasing order of $1$-shot "Mini" performance where available, and according to $5$-shot "Mini" otherwise}
\label{tab:semi}
\centering
\scriptsize{
\begin{tabular}[h]{lccccc}
\hline
& \# Unlabeled & Mini 1-shot & Mini 5-shot & Tiered 1-shot & Tiered 5-shot\\
\hline

\textbf{TPN \cite{Liu2019x}} & 360 &
52.78 $\pm$ 0.27 & 66.42 $\pm$ 0.21 & - & - \\

\textbf{PSN \cite{projective2019}} & 100 &
- & 68.12 $\pm$ 0.67 & - & 71.15 $\pm$ 0.67 \\

\textbf{TPN \cite{Liu2019x}} & 1170 &
 - & - & 55.74 $\pm$ 0.29 & 71.01 $\pm$ 0.23\\

\textbf{LST \cite{Li2019a}} & 30 &
65.00 $\pm$ 1.90 & - & 75.40 $\pm$ 1.60 & - \\

\textbf{SKM \cite{Ren2018}} & 100 &
62.10 $\pm$ N.A. & 73.60 $\pm$ N.A. & 68.60 $\pm$ N.A. & 81.00 $\pm$ N.A. \\

\textbf{TPN \cite{Liu2019x}} & 100 &
62.70 $\pm$ N.A. & 74.20 $\pm$ N.A. & 72.10 $\pm$ N.A. & 83.30 $\pm$ N.A. \\

\textbf{LST \cite{Li2019a}} & 50 &
- & 77.80 $\pm$ 0.80 & - & 83.40 $\pm$ 0.80 \\

\textbf{LST \cite{Li2019a}} & 100 &
70.10 $\pm$ 1.90 & 78.70 $\pm$ 0.80 & 77.70 $\pm$ 1.60 & 85.20 $\pm$ 0.80 \\

\hline

\textbf{ICA} & 30 &
72.00 $\pm$ 0.24 & 81.31 $\pm$ 0.16 & 80.24 $\pm$ 0.24 & 86.57 $\pm$ 0.17\\

\textbf{ICA} & 50 &
72.66 $\pm$ 0.24 & 81.96 $\pm$ 0.16 & 80.86 $\pm$ 0.24 & 87.03 $\pm$ 0.17\\

\textbf{ICA} & 100 &
72.80 $\pm$ 0.24 & 82.27 $\pm$ 0.16 & 80.91 $\pm$ 0.25 & 87.14 $\pm$ 0.17\\

\textbf{ICA + BKM} & 30 &
75.70 $\pm$ 0.22 & 83.59 $\pm$ 0.14 & 82.97 $\pm$ 0.23 & 88.34 $\pm$ 0.15\\

\textbf{ICA + BKM} & 50 &
76.46 $\pm$ 0.22 & 84.36 $\pm$ 0.14 & 83.51 $\pm$ 0.22 & 88.81 $\pm$ 0.15\\

\textbf{ICA + BKM} & 100 &
76.83 $\pm$ 0.22 & 84.83 $\pm$ 0.14 & 83.73 $\pm$ 0.22 & 88.95 $\pm$ 0.15\\

\textbf{ICA + MSP} & 30 &
78.55 $\pm$ 0.25 & 84.84 $\pm$ 0.14 & 85.04 $\pm$ 0.24 & 88.94 $\pm$ 0.15\\

\textbf{ICA + MSP} & 50 &
79.58 $\pm$ 0.25 & 85.41 $\pm$ 0.13 & 85.75 $\pm$ 0.24 & 89.32 $\pm$ 0.15\\

\textbf{ICA + MSP} & 100 &
80.11 $\pm$ 0.25 & 85.78 $\pm$ 0.13 & 86.00 $\pm$ 0.23 & 89.39 $\pm$ 0.15\\

\hline
\end{tabular}
}
\end{table}%

\vspace{-0.3cm}
\subsection{Ablation study}\label{sec:ablation}
\vspace{-0.2cm}
Here we describe the ablation experiments analyzing the different design choices and parameters of the proposed approaches, and of the problem setting itself.

\vspace{-0.2cm}
\subsubsection{Number of queries in transductive FSL.}\label{sec:num_queries}
\begin{figure}[!t]
\vspace{-0.9cm}
\centering
\subfloat[]{\includegraphics[width=6cm]{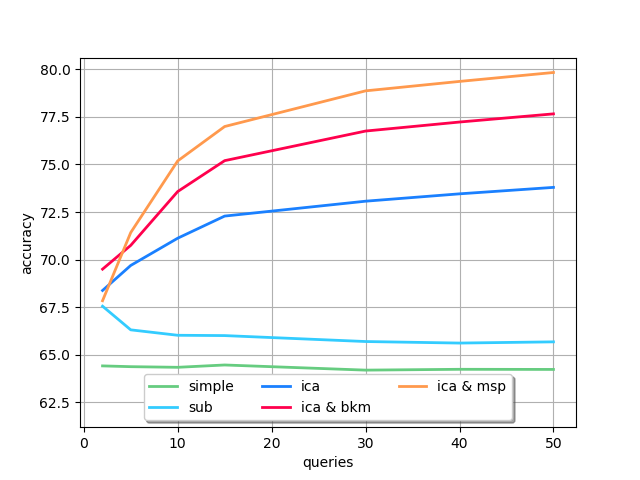} }
\subfloat[]{ \includegraphics[width=6cm]{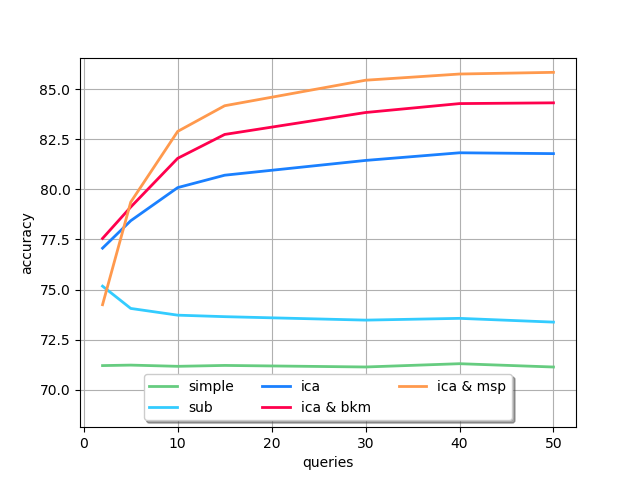} }
\vspace{-0.2cm}
\caption{{\bf Number of queries in transductive FSL setting:} (a) \textit{mini}ImageNet (Mini); (b) \textit{tiered}ImageNet (Tiered)  }
\label{fig:num_queries}
\end{figure}

\vspace{-0.2cm}
Since the unlabelled data in transductive FSL is comprised entirely from the query samples, the size of the query set $Q$ in the meta-testing episodes affects the performance. To test this we have evaluated the proposed TAFSSL ICA-based methods, as well as two baselines, namely SimpleShot \cite{Wang2019b}, and its adaptation to transductive setting "trans-mean-sub*" (sub). All methods were tested varying the number of queries from $2$ to $50$. The results of this ablation on both the Tiered and Mini benchmarks are shown on figure \ref{fig:num_queries}. As can be seen from the figure, already for as little as $5$ queries a substantial gap can be observed (for both the benchmarks) between the proposed best performing ICA+MSP technique and the best of the baselines.

\vspace{-0.2cm}
\subsubsection{Out of distribution noise (distraction classes) in unlabeled data.}\label{sec:noise}
\begin{figure}[!t]
\vspace{-0.9cm}
\centering
\subfloat[]{\includegraphics[width=6cm]{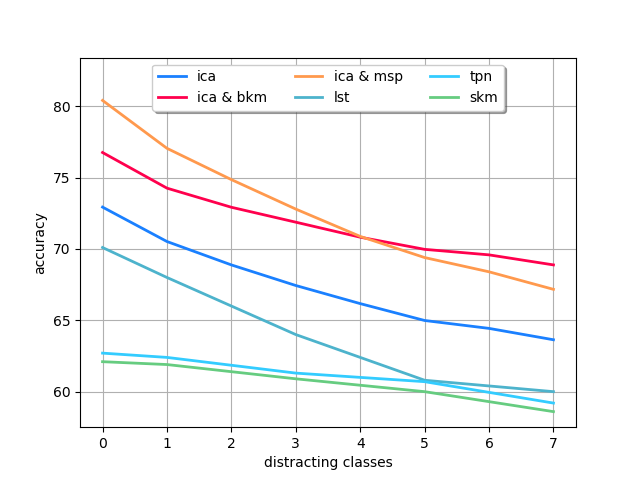} }
\subfloat[]{ \includegraphics[width=6cm]{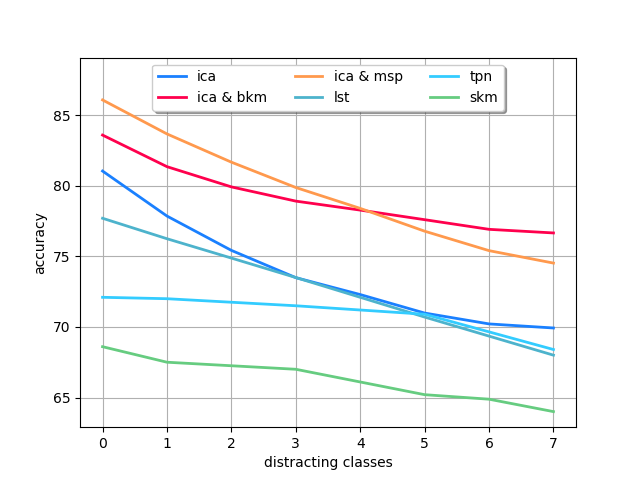} }
\vspace{-0.2cm}
\caption{{\bf Noise:} The figure shows the affect of the unlabeled data noise on the performance. Plots for LST \cite{Li2019a}, TPN \cite{Liu2019x}, and SKM \cite{Ren2018} are extrapolated from their original publications. (a) \textit{mini}ImageNet (Mini); (b) \textit{tiered}ImageNet (Tiered)  }
\label{fig:noise}
\end{figure}

In many applications, the unlabeled data may become contaminated with samples "unrelated" to the few-shot task $\mathcal{T}$ target classes distribution. This situation is most likely to arise in the semi-supervised FSL setting, as in transductive FSL the unlabeled samples are the queries and unless we are interested in open-set FSL mode (to the best of our knowledge not explored yet), these are commonly expected to belong only to the target classes distribution. In the semi-supervised FSL literature \cite{Ren2018,Liu2019x,Li2019a}, this type of noise is evaluated using additional random samples from random "distracting" classes added to the unlabeled set. In figure \ref{fig:noise} we compare our proposed ICA-based TAFSSL approaches to SOTA semi-supervised FSL methods \cite{Ren2018,Liu2019x,Li2019a}. By varying the number of distracting classes from $0$ to $7$, we see that about $8\%$ accuracy gap is maintained between top TAFSSL method and the top baseline across all the tested noise levels.

\vspace{-0.2cm}
\subsubsection{The number of TAFSSL sub-space dimensions.}\label{sec:dim}
\begin{figure}[!t]
\vspace{-0.7cm}
\centering
\subfloat[]{\includegraphics[width=6cm]{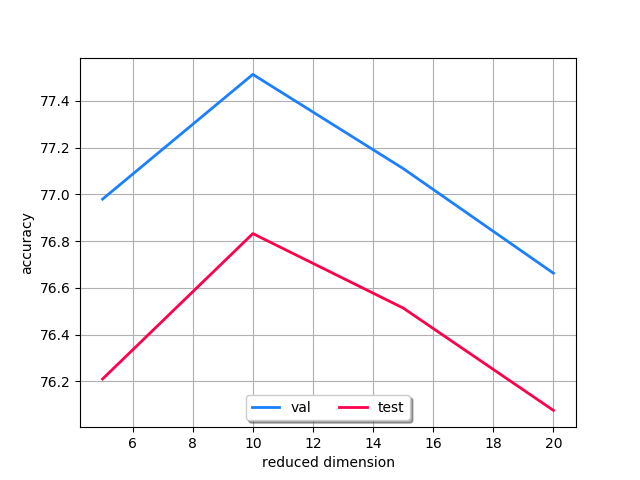} }
\subfloat[]{ \includegraphics[width=6cm]{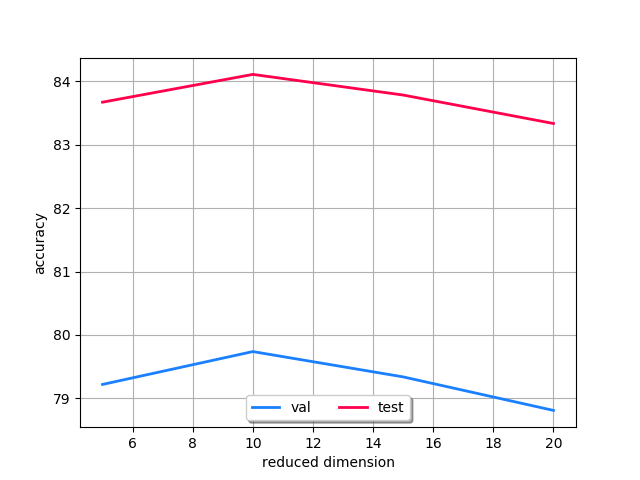} }
\caption{{\bf ICA dimension vs accuracy:} (a) \textit{mini}ImageNet (Mini) (b) \textit{tiered}ImageNet (Tiered)  }
\label{fig:dim}
\end{figure}

An important parameter for TAFSSL is the number of the dimensions of the sub-space selected by the TAFSSL approach. In figure \ref{fig:dim} we explore the effect of the number of chosen dimensions in ICA-based TAFFSL on both the Mini and the Tiered benchmarks. As can be seen from the figure, the optimal number of dimensions for ICA-based TAFSSL approaches is $10$, which is consistent between both test and validation sets. Interestingly, the same number $10$ is consistent between the two benchmarks. Similarly, using validation, the optimal dimension for PCA-based TAFSSL was found to be $4$ (also consistently on the two benchmarks).

\vspace{-0.2cm}
\subsubsection{Backbone architectures.}\label{sec:backbones}
The choice of backbone turned out to be an important factor for FSL methods performance \cite{closer_look,Wang2019b}. In Table \ref{tab:backbones} we evaluate the performance of one of the proposed TAFSSL approaches, namely PCA+MSP while using different backbones pre-trained on the training set to compute the base feature space $\mathcal{F}$. We used the $1$-shot transductive FSL setting on both Mini and Tiered benchmarks for this evaluation. As can be seen from the table, larger backbones produce better performance for the TAFSSL approach. In addition, we list the reported performance of the competing SOTA transductive FSL methods in the same table for direct comparison using the same backbones. As can be seen, above $8\%$ accuracy advantage is maintained by our proposed approach above the top previous method using the corresponding WRN architecture.
\begin{table}[!t]
\vspace{-0.5cm}
\centering
\caption{Backbones comparison. The $1$-shot transductive FSL setting for \textit{mini}ImageNet (Mini) and \textit{tiered}ImageNet (Tiered) was used for this comparison}
\label{tab:backbones}
\scriptsize{
\begin{tabular}[h]{lccc}
\hline
& Backbone & Mini 1-shot & Tiered 1-shot\\
\hline

\textbf{TPN \cite{Liu2019x}} & 
Conv-4 & 55.51 $\pm$ 0.86 & 59.91 $\pm$ 0.94\\

\textbf{TPN \cite{Liu2019x}} & 
ResNet-12 & 59.46 $\pm$ N.A. & -\\

\textbf{Transductive Fine-Tuning \cite{dhillon2019}} & 
WRN & 65.73 $\pm$ 0.68 & 73.34 $\pm$ 0.71 \\

\hline

\textbf{PCA + MSP} &
Conv-4 & 56.63 $\pm$ 0.27 & 60.27 $\pm$ 0.29\\

\textbf{PCA + MSP} &
ResNet-10 & 70.93 $\pm$ 0.28 & 76.27 $\pm$ 0.28\\

\textbf{PCA + MSP} &
ResNet-18 & 73.73 $\pm$ 0.27 & 80.60 $\pm$ 0.27\\

\textbf{PCA + MSP} &
WRN & 73.72 $\pm$ 0.27 & 81.61 $\pm$ 0.26\\

\textbf{PCA + MSP} &
DenseNet & 76.31 $\pm$ 0.26 & 84.06 $\pm$ 0.25\\

\hline
\end{tabular}
}
\end{table}

\vspace{-0.2cm}
\subsubsection{Unbalanced (long-tail) test classes distribution in unlabeled data.}\label{sec:unbalanced}
\begin{figure}[!t]
\vspace{-0.9cm}
\centering
\subfloat[]{\includegraphics[width=6cm]{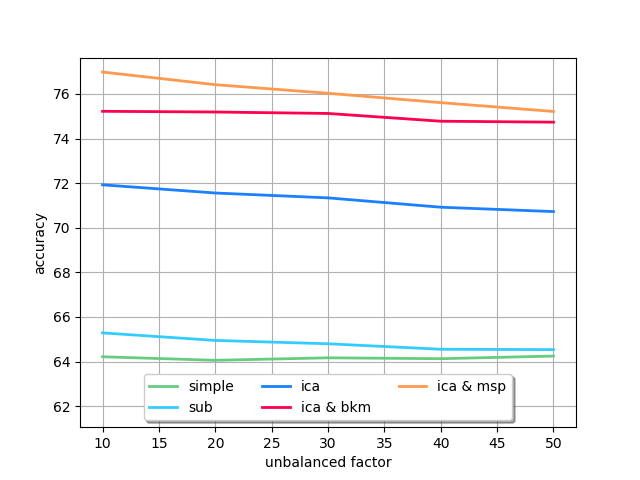} }
\subfloat[]{ \includegraphics[width=6cm]{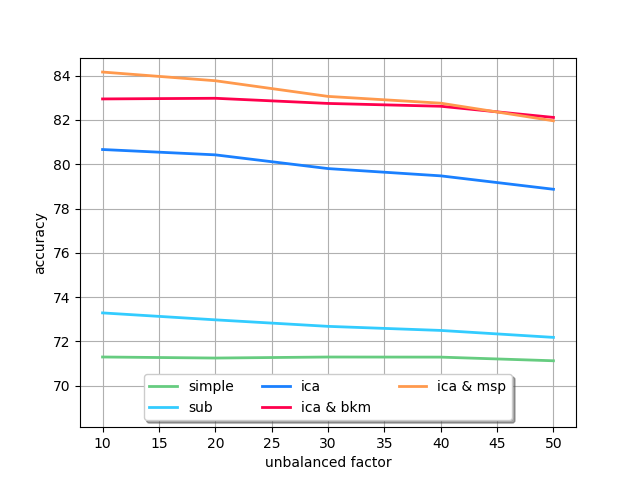} }
\vspace{-0.2cm}
\caption{{\bf Unbalanced:} (a) \textit{mini}ImageNet (Mini) (b) \textit{tiered}ImageNet (Tiered)}
\label{fig:unbalanced}
\end{figure}

In all previous transductive FSL works, the test tasks (episodes) were balanced in terms of the number of queries corresponding to each of the test classes. While this is fine for experimental evaluation purposes, in practical applications there is no guarantee that the bulk of queries sent for offline evaluation will be balanced in terms of classes. In fact, it is more likely that it will have some skew. To test the effect of query set skew (lack of balance) in terms of number of query samples per class, we have evaluated the proposed ICA-based TAFSSL approaches, as well as the Simple-Shot \cite{Wang2019b} 
and its transductive adaptation "trans-mean-sub*" (sub) baselines, under varying levels of query set skew. The level of skew was controlled through the so-called "unbalanced factor" 
parameter $R$: in each test episode, for each class $15 + uni([0,R])$ query samples were 
randomly chosen (here $uni$ refers to a uniform distribution). Figure \ref{fig:unbalanced} 
shows the effect of varying $R$ from $10$ to $50$, while at the extreme setting ($50$) above factor $4$ skew is possible between the classes in terms of the number of associated queries. Nevertheless, as can be seen from the figure, the effect of lack of balance on the performance of the TAFSSL based approaches is minimal, leading to at most $1\%$ performance loss at $R=50$. Since no prior work offered a similar evaluation design, we believe that the proposed protocol may become an additional important tool for evaluating transductive FSL methods under lack of query set balance in the future.
\vspace{-0.3cm}
\section{Summary and Conclusions}\label{sec:conclusion}
\vspace{-0.2cm}
In this paper we have highlighted an additional important factor on FSL classification performance - the Feature Sub-Space Learning (FSSL), and specifically it's Task Adaptive variant (TAFSSL). We have explored different methods and their combinations for benefiting from TAFSSL in few-shot classification and have shown great promise for this kind of techniques by achieving large margin improvements over transductive and semi-supervised FSL state-of-the-art, as well as over the more classical FSL that does not use additional unlabeled data, thus highlighting the benefit of the latter. Potential future work directions include incorporating TAFSSL into the meta-training (pre-training) process (e.g. by propagating training episodes gradients through pyTorch PCA/ICA implementations, and the proposed clustering techniques BKM/MSP); exploring non-linear TAFSSL variants (e.g. kernel TAFSSL, or using a small DNN); further exploring the effect of TAFSSL in \textit{any-shot} learning and the significance of the \textit{way} parameter of the task; exploring the benefits of TAFSSL in cross-domain few-shot learning where the FSL backbone pre-training occurs in different visual domain from the one test classes are sampled from.

%
%
\bibliographystyle{splncs04}
\bibliography{references_mendeley}
\end{document}